# Linear Segmentation and Segment Significance

## Min-Yen Kan*, Judith L. Klavans** and Kathleen R. McKeown*

Department of Computer Science* and Center for Research on Information Access**
Columbia University
New York, NY 10027, USA
{min,klavans,kathy}@cs.columbia.edu

## Abstract

We present a new method for discovering a segmental discourse structure of a document while categorizing each segment's function and importance. Segments are determined by a zero-sum weighting scheme, used on occurrences of noun phrases and pronominal forms retrieved from the document. Segment roles are then calculated from the distribution of the terms in the segment. Finally, we present results of evaluation in terms of precision and recall which surpass earlier approaches[1].

## Introduction

Identification of discourse structure can be extremely useful to natural language processing applications such as automatic text summarization or information retrieval (IR). For example, a summarization agent might chose to summarize each discourse segment separately. Also, segmentation of a document into blocks of topically similar text can assist a search engine in choosing to retrieve or highlight a segment in which a query term occurs. In this paper, we present a topical segmentation program that achieves a 10% increase in both precision and recall over comparable previous work.

In addition to segmenting, the system also labels the function of discovered discourse segments as to their relevance towards the whole. It identifies 1) segments that contribute some detail towards the main topic of the input, 2) segments that summarize the key points, and 3) segments that contain less important information. We evaluated our segment classification as part of a summarization system that utilizes highly pertinent segments to extract key sentences.

We investigated the applicability of this system on general domain news articles. Generally, we found that longer articles, usually beyond a three-page limit, tended to have their own prior segmentation markings consisting of headers or bullets, so these were excluded. We thus concentrated our work on a corpus of shorter articles, averaging roughly 800-1500 words in length: 15 from the *Wall Street Journal* in the Linguistic Data Consortium's 1988 collection, and 5 from the on-line *The Economist* from 1997. We constructed an evaluation standard from human segmentation judgments to test our output.

## 1 SEGMENTER: Linear Segmentation

For the purposes of discourse structure identification, we follow a formulation of the problem similar to Hearst (1994), in which zero or more segment boundaries are found at various paragraph separations, which identify one or more topical text segments. Our segmentation is linear, rather than hierarchical (Marcu 1997 and Yaari 1997), i.e. the input article is divided into a linear sequence of adjacent segments.

[1] This material is based upon work supported by the National Science Foundation under Grant No. (NSF #IRI-9618797) and by the Columbia University Center for Research on Information Access.

Our segmentation methodology has three distinct phases (Figure 1), which are executed sequentially. We will describe each of these phases in detail.

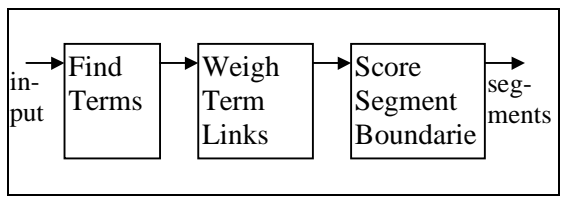

Figure 1. SEGMENTER Architecture

## 1.1 Extracting Useful Tokens

The task of determining segmentation breaks depends fundamentally on extracting useful topic information from the text. We extract three categories of information, which reflect the topical content of a text, to be referred to as *terms* for the remainder of the paper:

1. proper noun phrases;
2. common noun phrases;
3. personal and possessive pronouns.

In order to find these three types of terms, we first tag the text with part of speech (POS) information. Two methods were investigated for assigning POS tags to the text: 1) running a specialized tagging program or 2) using a simple POS table lookup. We chose to use the latter to assign tags for time efficiency reasons (since the segmentation task is often only a preprocessing stage), but optimized the POS table to favor high recall of the 3 term types, whenever possible[2]. The resulting system was faster than the initial prototype that used the former approach by more than a magnitude, with a slight decline in precision that was not statistically significant. However, if a large system requires accurate tags after segmentation and the cost of tagging is not an issue, then tagging should be used instead of lookup.

---

[2] We based our POS table lookup on NYU's COMLEX (Grishman et al. 1994). After simplifying COMLEX's categories to only reflect information important to to our three term types, we flattened all multi-category words (i.e. "jump" as V or N) to a single category by a strategy motivated to give high term recall (i.e. "jump" maps to N, because NP is a term type.)

Once POS tags have been assigned, we can retrieve occurrences of noun phrases by searching the document for this simple regular expression:

(Adj | Noun)* Noun

This expression captures a simple noun phrase without any complements. More complex noun phrases such as "proprietor of Stag's Leap Wine Cellars in Napa Valley" are captured as three different phrases: "proprietor", "Stag's Leap Wine Cellars" and "Napa Valley". We deliberately made the regular expression less powerful to capture as many noun phrases as possible, since the emphasis is on high NP recall.

After retrieving the terms, a post-processing phase combines related tokens together. For possessive pronouns, we merge each possessive with its appropriate personal pronoun ("my" or "mine" with "I", etc.) For noun phrases, we canonicalize noun phrases according to their heads. For example, if the noun phrases "red wine" and "wine" are found in a text, we subsume the occurrences of "red wine" into the occurrences of "wine", under the condition that there are no other "wine" headed phrases, such as "white wine".

Finally, we perform thresholding to filter irrelevant words, following the guidelines set out by Justeson and Katz (1995). We use a frequency threshold of two occurrences to determine topicality, and discard any pronouns or noun phrases that occur only once.

## 1.2 Weighting Term Occurrences

Once extracted, terms are then evaluated to arrive at segmentation.

### 1.2.1 Link Length

Given a single term (noun phrase or pronominal form) and the distribution of its occurrences, we link related occurrences together. We use proximity as our metric for relatedness. If two occurrences of a term occur within $n$ sentences, we link them together as a single unit, and repeat until no larger units can be built. This idea is a simpler interpretation of the notion of *lexical chains*. Morris and Hirst (1991) first proposed this notion to chain semantically related words together via a thesaurus, while we chose only repetition of the same stem word[3].

However, for these three categories of terms we noticed that the linking distance differs depending on the type of term in question, with proper nouns having the maximum allowable distance and the pronominal forms having the least. Proper nouns generally refer to the same entity, almost regardless of the number of intervening sentences. Common nouns often have a much shorter scope of reference, since a single token can be used to repeatedly refer to different instances of its class. Personal pronouns scope even more closely, as is expected of an anaphoric or referring expression where the referent can be, by definition, different over an active discourse. Any term occurrences that were not linked were then dropped from further consideration. Thus, link length or linking distance refers to the number of sentences allowed to intervene between two occurrences of a term.

### 1.2.2 Assigning Weights

After links are established, weighting is assigned. Since paragraph level boundaries are not considered in the previous step, we now label each paragraph with its positional relationship to each term's link(s). We describe these four categories for paragraph labeling and illustrate them in the figure below.

**Front**: a paragraph in which a link begins.
**During**: a paragraph in which a link occurs, but is not a front paragraph.
**Rear**: a paragraph in which a link *just stopped* occurring the paragraph *before*.
**No link**: any remaining paragraphs.

```
paras 1     2    3    4    5    6    7    8
sents 1234567890123456789012345678901234 5
wine :      1xx1                       1x21
type :n    f   d       r     n       f   d
```
Figure 2a. A term "wine", and its occurrences and type.

---

Figure 2a shows the algorithm as developed thus far in the paper, operating on the term "wine". The term appears a total of six times, as shown by the numbers in the central row. These occurrences have been grouped together into two term links, as joined by the "x"s. The bottom "type" line labels each paragraph with one of the four paragraph relations. We see that it is possible for a term to have multiple **front** or **rear** paragraphs, as illustrated, since a term's occurrences might be separated between disparate links.

Then, for each of the four categories of paragraph labeling mentioned before, and for each of the three term types, we assign a different segmentation score, listed in Table 1, whose values were derived by training, to be discussed in section 1.2.4.

| Term Type | Paragraph Type with respect to term | | | | Link Length |
|---|---|---|---|---|---|
| | front | rear | during | No link | |
| Proper NP | 10 | 8 | -3 | * | 8 |
| Common NP | 10 | 8 | -3 | * | 4 |
| Pronouns & Possessives | 1 | 13 | -1 | * | 0 |

Table 1 - Overview of weighting and linking scheme used in SEGMENTER; starred scores to be calculated later.

For noun phrases, we assume that the introduction of the term is a point at which a new topic may start; this is Youmans' (1991) Vocabulary Management Profile. Similarly, when a term is no longer being used, as in **rear** paragraphs, the topic may be closed. This observation may not be as direct as "vocabulary introduction", and thus presumably not as strong a marker of topic change as the former. Moreover, paragraphs in which the link persists throughout indicate that a topic continues; thus we see a negative score assigned to **during** paragraphs. When we apply the same paragraph labeling to pronoun forms, the same rationale applies with some modifications. Since the majority of pronoun referents occur before the pronoun (i.e. anaphoric as opposed to cataphoric), we do not weigh the **front** boundary heavily, but instead place the emphasis on the **rear**.

### 1.2.3. Zero Sum Normalization

When we iterate the weighting process described above over each term, and total the scores assigned, we come up with a numerical score for an indication of which paragraphs are more likely to beh a topical boundary. The higher the numerical score, the higher the likelihood that the paragraph is a beginning of a new topical segment. The question then is what should the threshold be?

```
paras 1    2    3      4      5        6    7    8
sents 1234567890123456789012345678901 2345
wine :     1xx1                          1x21
type :n    f    d      r      n        f    d
score:*    10  -3      8      *        10  -3

sum to balance in zero-sum weighting: +12
zero :-6   10  -3      8     -6        10  -3
```

Figure 2b. A term "wine", its links and score assignment to paragraphs.

To solve this problem, we zero-sum the weights for each individual term. To do this, we first sum the total of all scores assigned to any **front, rear** and **during** paragraphs that we have previously assigned a score to and then evenly distribute to the remaining **no link** paragraphs the negative of this sum. This ensures that the net sum of the weight assigned by the weighting of each term sums to zero, and thus the weighting of the entire article, also sums to zero. In cases where **no link** paragraphs do not exist for a term, we cannot perform zero-summing, and take the scores assigned as is, but this is in small minority of cases. This process of weighting followed by zero-summing is shown by the extending the "wine" example, in Figure 2b, as indicated by the `score` and `zero` lines.

With respect to individual paragraphs, the summed score results in a positive or negative total. A positive score indicates a boundary, i.e. the beginning of a new topical segment, whereas a negative score indicates the continuation of a segment. This use of zero sum weighting makes the problem of finding a threshold trivial, since the data is normalized around the value zero.

### 1.2.4 Finding Local Maxima

Examination of the output indicated that for long and medium length documents, zero-sum weighting would yield good results. However, for the documents we investigated, namely documents of short length (800-1500 words), we have observed that multiple consecutive paragraphs, all with a positive summed score, actually only have a single, true boundary. In these cases, we take the maximal valued paragraph for each of these clusters of positive valued paragraphs as the only segment boundary. Again, this only makes sense for paragraphs of short length, where the distribution of words would smear the segmentation values across paragraphs. In longer length documents, we do not expect this phenomenon to occur, and thus this process can be skipped. After finding local maxima, we arrive at the finalized segment boundaries.

### 1.3 Algorithm Training

To come up with the weights used in the segmentation algorithm and to establish the position criteria used later in the segment relevance calculations, we split our corpus of articles in four sets and performed 4-fold cross validation training, intentionally keeping the five *Economist* articles together in one set to check for domain specificity. Our training phase consisted of running the algorithm with a range of different parameter settings to determine the optimal settings. We tried a total of 5 x 5 x 3 x 3 = 225 group settings for the four variables (**front**, **rear**, **during** weights and linking length settings) for each of the three (common nouns, proper nouns and pronoun forms) term types. The results of each run were compared against a standard of user segmentation judgments, further discussed in Section 3.

The results noted that a sizable group of settings (approximately 10%) seemed to produce very close to optimal results. This group of settings was identical across all four cross validation training runs, so we believe the algorithm is fairly robust, but we cannot safely conclude this without constructing a more extensive training/testing corpus.

## 2    SEGNIFIER: Segment Significance

Once segments have been determined, how can we go about using them? As illustrated in the introduction, segments can be utilized "as-is" by information retrieval and automatic summarization applications by treating segments as individual documents. However, this approach loses information about the cohesiveness of the text as a whole unit. What we are searching for is a framework for processing segments both as 1) sub-documents of a whole, and as 2) independent entities. This enables us to ask a parallel set of general questions concerning 1) how segments differ from each other, and 2) how a segment contributes to the document as a whole.

In this portion of the paper, we deal with instances of the two questions: 1) Can we decide whether a text segment is *important*? 2) How do we decide what type of *function* a segment serves? These two questions are related; together, they might be said to define the task of finding *segment significance*. We will show a two-stage, sequential approach that attempts this task in the context of the article itself. Assessing segment significance with respect to a specific query could be quite different.

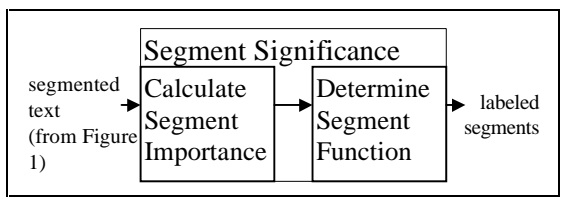

Figure 3 - SEGNIFIER Architecture

## 2.1 Segment Importance

Informally, segment importance is defined as the degree to which a given segment presents key information about the article as a whole. Our method for calculating this metric is given in the section below.

We apply a variant of Salton's (1989) information retrieval metric, *Term Frequency \* Inverse Document Frequency* (TF\*IDF) to noun phrases (no pronominial tokens are used in this algorithm). Intuitively, a segment containing noun phrases which are then also used in other segments of the document will be more central to the text than a segment that contains noun phrases that are used only within that one segment. We call this metric TF\*SF[4], since we base the importance of a segment on the distribution of noun phrases within the document. Note that this is not exactly analogous to IDF; we do not compute inverse segment frequency (ISF); this is because we are looking for segments with noun phrases that occur throughout a text rather that segments which are characterized by local noun phrases. Higher scores along the TF\*SF metric indicate a more central segment, which we equate with segment importance. SEGNIFIER first calculates the TF\*SF score for each noun phrase term using the term occurrence information and segment boundaries provided by the segmentation program.

---

[4] SF = Segment frequency (How many segments does the term occur in)

However, segment importance cannot be derived from merely summing together each term's TF\*SF score; we must also track in which segments the noun phrase occurs. This is needed to decide the coverage of the noun phrase in the segment. We illustrate segment coverage by the example of two hypothetical segments A-2 and B-2 in Figure 4. If we assert that the terms in each segment are equivalent, we can show that segment B-2 has better coverage because two noun phrases in B-2 taken together appear across all three segments, whereas in A-2 the noun phrase cover only two segments.

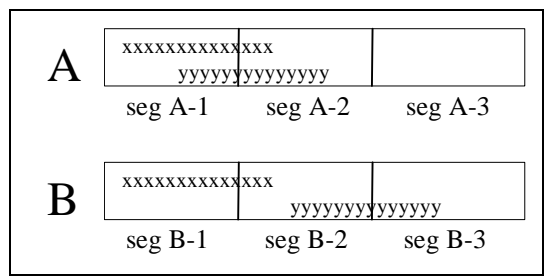

Figure 4 - Segment NP Coverage

To calculate coverage, SEGNIFIER first iterates over all the occurrences of all terms within a segment, and then increments the score. The increment depends on the number of terms previously seen that also fall in the same segment. We use a harmonic series to determine the score: for the first occurrence of a term in some segment, 1 is added to the segment's coverage score; a second occurrence adds 1/2; a third, 1/3, and so forth.

We normalize both the sum of the TF*SF scores over its terms and its coverage score to calculate the segment importance of a segment. Segment importance in our current system is given by a sum of these two numbers; thus the range is from 0.0 (not important) to 2.0 (maximally important). We summarize the algorithm for calculating segment importance in the psuedocode in Figure 5 below.

```
for each segment {
  { // TF*SF calculation
    TF_SF = sum of TF_SF per NP term;
    TF_SF = TF_SF /
         (max TF_SF over all segments);
  }
  { // coverage calculations
    coverage = sum of coverage per NP term;
    coverage = coverage /
         (max coverage over all segments);
  }
  seg_importance = TF_SF + coverage;
}
```

Figure 5 - Segment importance psuedocode

## 2.2 Segment Functions

Contrasting with segment importance, which examines the prominence of a segment versus every other segment, we now turn to examine segment function, which looks at the role of the segment in discourse structure. We currently classify segments into three types:

**a. Summary Segments** – A summary segment contains a summary of the article. We assume either the segment functions as an overview (towards the beginning of an article) or as a conclusion (near the end of an article), so the position of the segment within the document is one of the determining factors. According to our empirical study, summary segments are segments with the highest segment importance out of segments that occur within the first and last 20% of an article. In addition, the importance rating must be among the highest 10% of *all* segments.

**b. Anecdotal Segments** – Material that draws a reader into the main body of the article itself are known in the field of journalism as *anecdotal leads*. Similarly, closing remarks are often clever comments for effect, but do not convey much content. In our attempts to try to detect these segments, we have restricted our scope to the first and last segments of an article.

Empirical evidence suggests that in the domain of journalistic text, at least a single person is introduced during an anecdotal segment, to relate the interesting fact or narrative. This person is often not mentioned outside the segment; since the purpose of relating the anecdote is limited in scope to that segment. Accordingly, SEGNIFIER looks for a proper noun phrase that occurs only within the candidate segment, and not in other segments. This first or last segment is then labeled as **anecdotal**, if it has not been already selected as the summary segment. This method worked remarkably well on our data although we need to address cases where the anecdotal material has a more complex nature. For example, anecdotal material is also sometimes woven throughout the texts of some documents.

**c. Support Segments** – These segments are the default segment type. Currently, if we cannot assign a segment as either a summary or an anecdotal segment, it is deemed to be a support segment.

## 2.3 Related work on Segment Significance

There has been a large body of work done of assessing the importance of passages and the assignment of discourse functions to them. Chen and Withgott (1992) examine the problem of audio summarization in domain of speech, using instances emphasized speech to determine and demarcate important phrases. Although their work is similar to the use of terms to demarcate segments, the nature of the problem is different. The frequency of terms in text versus emphasized speech in audio forces different approaches to be taken. Singhal and Salton (1996) examined determining paragraph connectedness via vector space model similarity metrics, and this approach may extend well to the segment level. Considering the problem from another angle, discourse approaches have focused on shorter units than multi-paragraph segments, but Rhetorical Structure Theory (Marcu 1997 and others) may be able to scale up to associate rhetorical functions with segments. Our work is a first attempt to bring these fields together to solve the problem of segment importance and function.

| | 15 WSJ | | | | 5 Economist | | | | Total | | | |
|---|---|---|---|---|---|---|---|---|---|---|---|---|
| | Precision | | Recall | | Precision | | Recall | | **Precision** | | **Recall** | |
| | avg | S.D. | avg | S.D. | avg | S.D. | avg | S.D. | **avg** | **S.D.** | **avg** | **S.D.** |
| Monte Carlo 33% | 29.0% | 9.2 | 33.3% | .02 | 32.8% | 12.6 | 33.3% | .02 | 29.8% | 9.9 | 33.3% | .02 |
| Hypergeometric | 30.6% | N/A | 30.6% | N/A | 32.9% | N/A | 32.9% | N/A | 32.0% | N/A | 32.0% | N/A |
| TEXTTILING | 28.2% | 18.1 | 33.4% | 25.9 | 18.3% | 20.7 | 18.7% | 18.5 | 25.8% | 18.7 | 29.8% | 27.8 |
| SEGMENTER | 47.0% | 21.4 | 45.1% | 24.4 | 28.6% | 26.2 | 22.67% | 25.2 | 42.6% | 23.5 | 39.6% | 25.9 |
| Human Judges | 67.0% | 11.4 | 80.4% | 8.9 | 55.8% | 17.2 | 71.9% | 4.6 | 62.4% | 13.5 | 78.2% | 87.6 |

Table 2 - Evaluation Results on Precision and Recall Scales

# 3    Evaluation

## 3.1    Segmentation Evaluation

For the segmentation algorithm we used a web-based segmentation evaluation facility to gather segmentation judgments. Each of the 20 articles in the corpus was segmented by at least four human judges, and the majority opinion of segment boundaries was computed as the evaluation standard  (Klavans et al. 1998).

Human judges achieved on average only 62.4% agreement with the majority opinion, as seen in Table 2. Passonneau and Litman (1993) show that this surprisingly low agreement is often the result of evaluators being divided between those who regard segments as more localized and those who prefer to split only on large boundaries.

We then verified that the task was well defined by testing for a strong correlation between the markings of the human judges. We test for inter-judge reliability using Cochran (1950)'s *Q-test*, also discussed in Passonneau and Litman (1993). We found a very high correlation between judges indicating that modeling the task was indeed feasible; the results showed that there was less than a 0.15% chance on average that the judges' segment marks agreed by chance. We also calculated Kappa (*K*), another correlation statistic that corrects for random chance agreement. Kappa values range from -1.0, showing complete negative correlation to +1.0, indicating complete positive correlation. Surprisingly, the calculations of *K* showed only a weak level of agreement between judges (*K* avg = .331, S.D.= .153). Calculations of the significance of *K* showed that results were generally significant to the 5% level, indicating that although the interjudge agreement is weak, it is statistically significant and observable.

We computed SEGMENTER's performance by completing the 4-fold cross validation on the test cases. Examining SEGMENTER's results show a significant improvement over the initial algorithm of Hearst 1994 (called TEXTTILING), both in precision and recall. A future step could be to compare our segmenting algorithm against other more recent systems (such as Yaari 1997, Okumura and Honda 1994).

We present two different baselines to compare the work against. First, we applied a Monte Carlo simulation that segments at paragraph breaks with a 33% probability. We executed this baseline 10,000 times on each article and averaged the scores. A more informed baseline is produced by applying a hypergeometric distribution, which calculates the probability of some number of successes by sampling without replacement. For example, this distribution gives the expected number of red balls drawn from a sample of *n* balls from an urn containing *N* total balls, where only *r* are red. If we allow the number of segments, *r*, to be given, we can apply this to segmentation to pick *r* segments from *N* paragraphs. By comparing the results in Table 3, we can see that the correct number of segments (*r*) is difficult to determine. TEXTTILING's performance falls below the hypergeomtric baseline, but on the average, SEGMENTER outperforms it.

However, notice that the performance of the algorithm and TEXTTILING quoted in this paper are low in comparison to reports by others. We believe this is due to the weak level of agreement between judges in our training/testing evaluation corpus. The wide range of performance hints at the variation which segmentation algorithms may experience when faced with different kinds of input.

## 3.2. Segment Significance Evaluation

As mentioned previously, segments and segment type assessments have been integrated into a key sentence extraction program (Klavans et al. 1998). This summary-directed sentence extraction differs from similar systems in its focus on high recall; further processing of the retrieved sentences would discard unimportant sentences and clauses. This system used the location of the first sentence of the **summary segment** as one input feature for deciding key sentences, along with standard features such as title words, TF*IDF weights for the words of a sentence, and the occurrences of communication verbs. This task-based evaluation of both modules together showed that combining segmentation information yielded markedly better results. In some instances only segmentation was able to identify certain key sentences; all other features failed to find these sentences. Overall, a 3.1% improvement in recall was directly achieved by adding segment significance output, increasing the system's recall from 39% to 42%. Since the system was not built with precision as a priority, so although precision of the system dropped 3%, we believe the overall effects of adding the segmentation information was valuable.

## 4    Future Work

Improvements to the current system can be categorized along the lines of the two modules. For *segmentation*, applying machine learning techniques (Beeferman et al. 1997) to learn weights is a high priority. Moreover we feel shared resources for segmentation evaluation should be established[5], to aid in a comprehensive cross-method study and to help alleviate the problems of significance of small-scale evaluations as discussed in Klavans et al (1998).

For judging *segment function*, we plan to perform a direct assessment of the accuracy of segment classification. We want to expand and refine our definition of the types of segment function to include more distinctions, such as the difference between document/segment borders (Reynar 1994). This would help in situations where input consists of multiple articles or a continuous stream, as in Kanade et al. (1997).

## 5    Conclusion

In this paper we have shown how multi-paragraph text segmentation can model discourse structure by addressing the dual problems of computing topical text segments and subsequently assessing their significance. We have demonstrated a new algorithm that performs linear topical segmentation in an efficient manner that is based on linguistic principles. We achieve a 10% increase in accuracy and recall levels over prior work (Hearst 1994, 1997). Our evaluation corpus exhibited a weak level of agreement among judges, which we believe correlates with the low level of performance of automatic segmentation programs as compared to earlier published works (Hearst 1997).

Additionally, we describe an original method to evaluate a segment's significance: a two part metric that combines a measure of a segment's generality based on statistical approaches, and a classification of a segment's function based on empirical observations. An evaluation of this metric established its utility as a means of extracting key sentences for summarization.

## Acknowledgements

The authors would like to thank Slava M. Katz, whose insights and insistence on quality work helped push the first part of the research forward. We are also indebted to Susan Lee of the University of California, Berkeley[6] for providing empirical validation of the segment significance through her key sentence extraction system. Thanks are also due to Yaakov Yaari of Bar-Ilan University, for helping us hunt down additional segmentation corpora. Finally, we thank the anonymous reviewers and the Columbia natural language group members, whose careful critiques led to a more careful evaluation of the paper's techniques.

---

[5] For the purposes of our own evaluation, we constructed web-based software tool that allows users to annotate a document with segmentation markings. We propose initiating a distributed cross evaluation of text segmentation work, using our system as a component to store and share user-given and automatic markings.

---

[6] Who was supported by the Computing Research Association.